\documentclass[letterpaper, 10 pt, conference]{ieeeconf}  

\IEEEoverridecommandlockouts                              

\usepackage{graphicx}
\usepackage{booktabs}
\usepackage{algorithmic}
\usepackage{algorithm}
\usepackage{amsmath,amssymb}
\usepackage{multirow}
\usepackage{makecell} 
\usepackage{amsmath}
\usepackage{comment}
\usepackage{xcolor}
\usepackage{colortbl}
\newcommand{\myparagraph}[1]{%
  \noindent\textbf{#1}%
}

\usepackage{amsmath}
\usepackage{bm}
\usepackage{url}
\usepackage{caption}
\usepackage{hyperref}
\usepackage{epigraph}

\title{\LARGE \bf
Developing Vision-Language-Action Model from Egocentric Videos
}

\author{Tomoya Yoshida$^{1,4}$, Shuhei Kurita$^{2,3,4}$, Taichi Nishimura$^{5}$, Shinsuke Mori$^{1}$
\thanks{$^{1}$Kyoto University, Kyoto 606-8501, Japan}%
\thanks{$^{2}$National Institute of Informatics, Tokyo 101-8430, Japan}%
\thanks{$^{3}$Institute of Science Tokyo,
Tokyo, Japan, Tokyo 152-8550, Japan}%
\thanks{$^{4}$NII LLMC, Tokyo 100-0003, Japan }%
\thanks{$^{5}$Sony Interactive Entertainment, Tokyo 108-0075, Japan}
\thanks{Contact: \href{mailto:yoshida.tomoya.25h@st.kyoto-u.ac.jp}{\nolinkurl{yoshida.tomoya.25h@st.kyoto-u.ac.jp}}}
}%

\begin{document}

\maketitle
\thispagestyle{empty}
\pagestyle{empty}


\begin{abstract}
Egocentric videos capture how humans manipulate objects and tools, providing diverse motion cues for learning object manipulation. Unlike the costly, expert-driven manual teleoperation commonly used in training Vision-Language-Action models (VLAs), egocentric videos offer a scalable alternative. However, prior studies that leverage such videos for training robot policies typically rely on auxiliary annotations, such as detailed hand-pose recordings. Consequently, it remains unclear whether VLAs can be trained directly from raw egocentric videos.
In this work, we address this challenge by leveraging EgoScaler, a framework that extracts 6DoF object manipulation trajectories from egocentric videos without requiring auxiliary recordings. We apply EgoScaler to four large-scale egocentric video datasets and automatically refine noisy or incomplete trajectories, thereby constructing a new large-scale dataset for VLA pre-training.
Our experiments with a state-of-the-art $\pi_0$ architecture in both simulated and real-robot environments yield three key findings: (i) pre-training on our dataset improves task success rates by over 20\% compared to training from scratch, (ii) the performance is competitive with that achieved using real-robot datasets, and (iii) combining our dataset with real-robot data yields further improvements.
These results demonstrate that egocentric videos constitute a promising and scalable resource for advancing VLA research.
\end{abstract}
\section{Introduction}

\begin{figure*}
    \centering
    \includegraphics[width=0.96\linewidth]{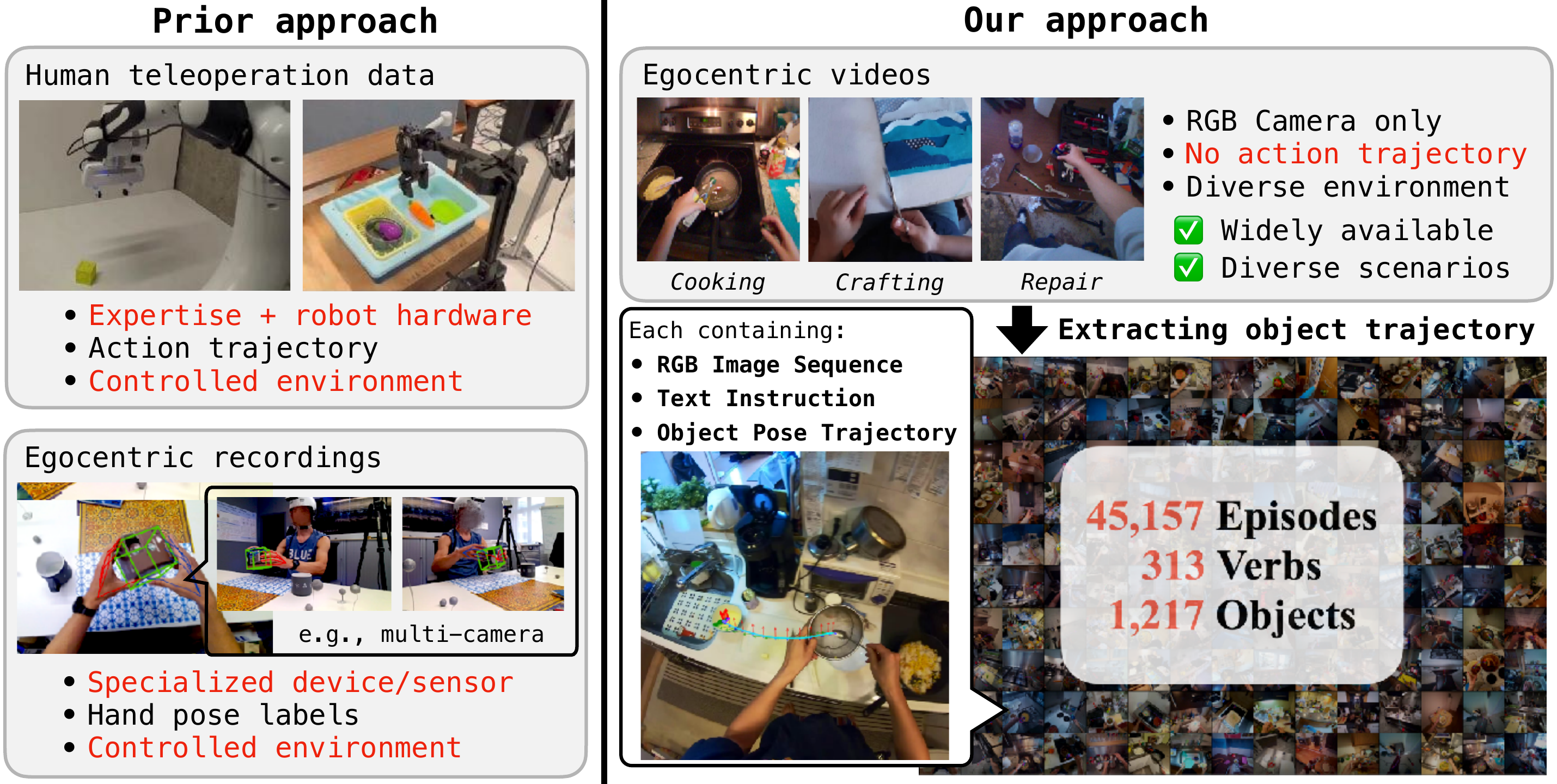}
    \vspace*{-3pt}
    \caption{\textbf{Comparison of conventional VLA pre-training sources and ours.}
    We leverage egocentric videos \textit{without auxiliary labels} for VLA pre-training.
    Using EgoScaler~\cite{egoscaler} to extract 6DoF object manipulation trajectories, we construct a large-scale dataset. The multi-camera example is adapted from \cite{h2o}.
    }
    \label{fig:teaser}
\end{figure*}

Vision-Language-Action models (VLAs) aim to learn general-purpose robot behaviors that follow natural language instructions across environments~\cite{rt-1, rt-2, oxe, octo, openvla, pi0, smolvla, agibot, gr00t}.
Such models are pre-trained with large-scale, multi-embodiment datasets~\cite{oxe, pi0, gr00t} and then fine-tuned on embodiment-specific datasets.
However, most pre-training datasets for VLAs heavily rely on human teleoperation, where a number of experts directly manipulate robots to collect instances for imitation learning.
This is inherently costly and labor-intensive, leaving a data scarcity problem.

One promising direction to address this problem is to leverage first-person perspective recordings of humans performing everyday tasks, enabled by the advancement of AR/VR devices and smart glasses~\cite{quest3, aria-glasses, vision-pro}.
Particularly, such egocentric videos provide diverse human-object interactions at a close range and inherently provide motion cues for learning object manipulation.
Several studies have begun to explore how to utilize egocentric videos in robot learning~\cite{r3m, egomimic, egovla, egodex}.
For example, EgoMimic~\cite{egomimic} and EgoVLA~\cite{egovla} leverage enriched egocentric recordings including hand poses to learn robot policies.
These studies demonstrate that utilizing egocentric videos is more time- and scale-efficient than those from teleoperation-based data collection.
Unfortunately, these approaches depend on dense auxiliary recordings, such as hand poses and action start/end timestamps.
Obtaining these dense auxiliary recordings requires specialized hardware, such as multi-camera systems or depth sensors, as well as extensive manual annotation.
In a recent study, LAPA~\cite{lapa} attempted to learn latent action representations from egocentric videos. 
While this approach is scalable because it does not require auxiliary labels, such latent representations often struggle to capture fine-grained motions. For example, they showed strong performance on simple actions like pushing but only mediocre performance on more complex skills like pick-and-place.

Considering the limited scalability of rich egocentric recordings and the lack of fine-grained motion cues in egocentric videos, existing methods provide valuable insights but may fall short of offering sufficiently detailed and diverse action examples for robotic foundation models (see Fig. \ref{fig:teaser}).
It is also notable that robot policies trained on diverse real-world egocentric recordings can fall short when evaluated within controlled environments, particularly simulators, due to simplified visual systems~\cite{simplerenv, sim2real-survey}. 
Therefore, although egocentric recordings are undeniably valuable resources of human motion cues, they remain underexplored in the existing literature.

To address this issue, we focus on extracting explicit action trajectories, which provide supervision that represents how to move and rotate objects.
We leverage EgoScaler~\cite{egoscaler}, a framework designed to extract object manipulation trajectories from egocentric videos.  
Each pose in a trajectory represents the centroid and rotation of the manipulated object, approximated as the end-effector states of a robot, excluding the gripper.
We apply this framework to four large egocentric video datasets, including Ego4D~\cite{ego4d}, Ego-Exo4D~\cite{egoexo4d}, HD-EPIC~\cite{hd-epic}, and Nymeria~\cite{nymeria}.
The extracted trajectories are then curated by automatically removing noisy or incomplete instances. After this careful filtering process, we construct a new large-scale dataset for VLA pre-training.

We conduct our experiments based on a state-of-the-art $\pi_0$~\cite{pi0} architecture.
For comparison, we include three real-robot datasets—BC-Z~\cite{bc-z}, BridgeData V2~\cite{bridge}, and Fractal~\cite{rt-1}, which match our dataset in scale and diversity.
We evaluate performance in both simulated (SIMPLER~\cite{simplerenv}) and real-robot (ALOHA~\cite{aloha}) environments.
Our key findings are threefold:
\begin{enumerate}
    \item We successfully train $\pi_0$ from egocentric videos without auxiliary labels, achieving significant improvements over both training from scratch and LAPA.
    \item Our dataset achieves performance on par with leading real-robot datasets, while slightly outperforming BC-Z and BridgeData V2.
    \item Combining our dataset with BridgeData V2 yields further gains, surpassing the performance of pre-training on either dataset alone.
\end{enumerate}
\section{Related Work}
\subsection{Dataset for Robotic Foundation Model}
Robotic foundation models have been revolutionizing the robot learning literature, as they enable the execution of multi-purpose and environment-agnostic actions across diverse robotic hardware.
Such robotic foundation models, however, rely on large-scale datasets for robotic actions that typically cover multiple robotic hardware and diverse tasks~\cite{oxe, pi0, gr00t, agibot}. 
As constructing such collections requires substantial effort from the robotics community, several datasets have been created through worldwide collaborations~\cite{droid,oxe}. Open X-Embodiment (OXE) dataset~\cite{oxe} aggregates over 60 individual datasets from various institutions into a unified collection that spans diverse manipulation tasks across multiple embodiments.
In OXE, they demonstrate that scaling datasets is crucial for improving model performance.
Similarly, $\pi_0$ model~\cite{pi0} depends on the internal dataset of `$\pi$,' which consists of over 10,000 hours of dexterous manipulation data and makes it possible to execute environment-agnostic and long-horizon tasks.
While these datasets have substantially advanced robot learning, the reliance on extensive human effort to construct large-scale robot datasets has become a major bottleneck to progress toward a general-purpose robotic foundation model.
To address this inherent limitation of manual dataset collection, we leverage egocentric videos to automatically construct a pre-training dataset for robotic foundation models.

\subsection{Egocentric Video Dataset}
Egocentric vision captures fine-grained hand–object interactions, providing motion cues for action understanding~\cite{h2o, hoi4d}. 
Reflecting its importance, a number of datasets have been introduced over the past decade~\cite{ego4d, egoexo4d, epic-kitchens55, epic-kitchens100, assembly101, egtea, hd-epic, egolife}. For example, Ego4D~\cite{ego4d} comprises over 3,000 hours of human daily activity videos spanning hundreds of scenarios, including cooking and crafting. 
Although egocentric videos remain relatively limited in scale compared to internet videos, the domain is expected to grow with the advancement of AR/VR agents and smart glasses~\cite{outlook-ego-future, egolife, holoassist}. 
This anticipated growth will further expand egocentric video collections, underscoring their potential as a scalable resource for robot learning.

\begin{figure*}
    \centering
    \includegraphics[width=0.94\linewidth]{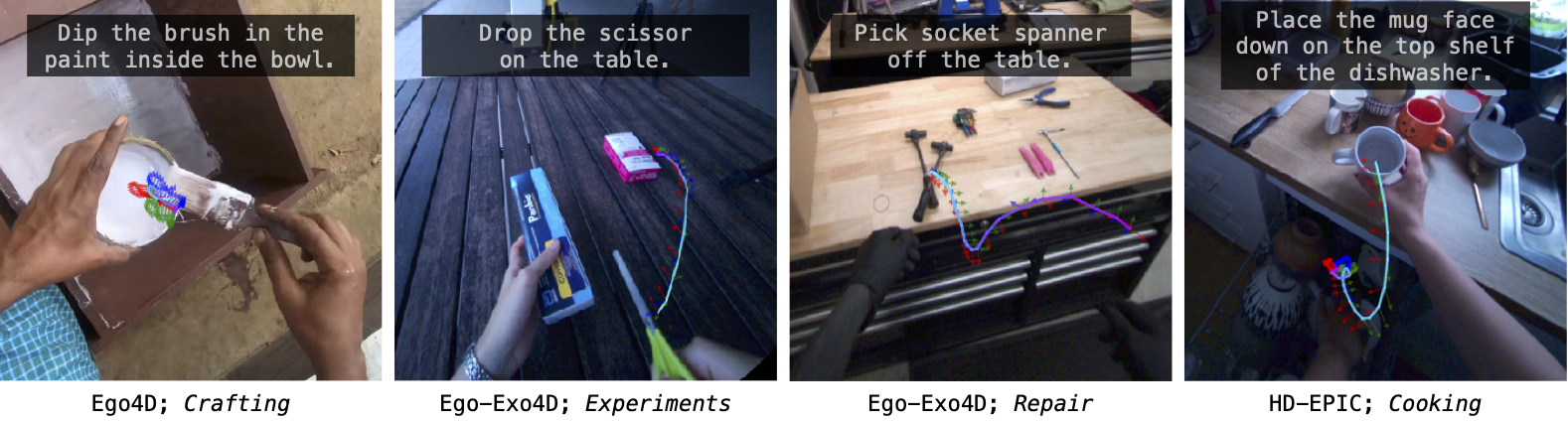}
    \vspace*{-4pt}
    \caption{\textbf{Samples of extracted trajectories via EgoScaler~\cite{egoscaler}.} The trajectory is color-coded from \textcolor{cyan}{cyan} (start) to \textcolor{purple}{purple} (end) to indicate temporal progression. Red, green, and blue arrows represent the X, Y, and Z axes of the object's coordinate frame at each time step.
    }
    \label{fig:egoscaler-example}
\end{figure*}

\subsection{Robot Learning from Human Demonstration Recordings}

Recordings of human everyday tasks in the real world have emerged as a promising source for robot learning, reducing reliance on costly teleoperation-based data collection~\cite{r3m, lapa, uniskill}.
In VRB~\cite{vrb} and HRP~\cite{hrp}, they extract visual affordances during human-object interaction for grasping tasks from egocentric videos to predict potential interaction regions, providing implicit guidance for robotic dexterous manipulation tasks.
Beyond grasping capabilities, a variety of approaches have been explored for more complex manipulation tasks~\cite{egomimic, mimicplay, phantom, egodex, zeromimic, vidbot, egovla, egozero, masquerade}.
MimicPlay~\cite{mimicplay} introduces an approach to learn high-level action plans by imitating hand trajectories from multi-view recordings of human demonstrations. 
By incorporating these high-level plans as latent representations into robot policies, this work achieves improved performance with minimal robot data.
More recently, EgoVLA~\cite{egovla} pre-trains on egocentric recordings including hand poses captured with specialized hardware, significantly enhancing VLA performance compared to training from scratch.
Although the utilization of egovision in robot learning seems successful, these existing approaches rely on recordings captured with specialized hardware, such as multi-camera systems, depth sensors, or proprietary devices like Aria Glasses~\cite{aria-glasses}, which limits the scalability of the resulting datasets. In contrast, our work focuses on constructing a pre-training dataset from egocentric videos without auxiliary labels, thereby unlocking the scalability of large-scale egocentric video resources for robot learning.
\section{Method}
Considering the cost and scalability issues of collecting teleoperation data, we leverage egocentric videos for VLA pre-training.
However, egocentric videos do not directly provide action trajectories.
To address this, we use EgoScaler~\cite{egoscaler}, a framework that extracts object manipulation trajectories from egocentric videos.
Applying this framework to large and diverse egocentric video datasets, we construct a large-scale pre-training dataset.
As shown in Fig.~\ref{fig:teaser}, each episode in this dataset comprises an image sequence, a text instruction, and a 6DoF object pose trajectory. 
We pre-train a VLA on this dataset and then post-train it on small embodiment-specific datasets.

\subsection{Problem Definition}
Following previous studies~\cite{aloha,pi0,pi0fast}, we train a robot policy that outputs sequences of future robot actions (action chunks).
Formally, at each timestep \(t\), given a language instruction \(\ell\), RGB observations \(v_t\), and proprioceptive state \(\boldsymbol{\tau}_t\), we model a policy 
\(\pi(\mathbf{a}_{1:H}\mid v_t,\boldsymbol{\tau}_t,\ell)\) that defines a distribution over the next \(H\) actions \(\{\mathbf{a}_t,\dots,\mathbf{a}_{t+H-1}\}\).   
In pre-training, \(v_t\) is a single image, and both \(\boldsymbol{\tau}_t\) and \(\mathbf{a}_t\) are approximated in an end-effector space without gripper states, derived from egocentric videos. 
In post-training and evaluation, \(v_t\) may comprise images from one or several cameras, depending on the hardware design, and both \(\boldsymbol{\tau}_t\) and \(\mathbf{a}_t\) are defined in the robot’s native control space (e.g., joint space or end-effector space).
Despite these gaps, recent studies have demonstrated that VLAs are capable of dealing with such modality differences~\cite{oxe,pi0}.
At inference time, an action chunk $a_{1:H} \sim \pi(a_{1:H}\mid v_t,\tau_t,\ell)$ is sampled from the trained policy and then executed sequentially.

\subsection{Pre-training Dataset Construction}
\label{subsec:dataset-construction}

\myparagraph{Re-visiting EgoScaler Framework.}
EgoScaler~\cite{egoscaler} extracts a 6DoF object manipulation trajectory from an egocentric video. This framework consists of four stages.
First, given a video clip, the start and end timestamps of the action as well as the manipulated object within the scene are identified using GPT-4o~\cite{gpt4o}.
Second, the position sequence of the manipulated object is extracted using an open‑vocabulary segmentation model~\cite{grounding-dino, sam} and a dense 3D point tracker~\cite{spatracker}. 
Third, this sequence is projected into the camera coordinate system of the action start frame via point cloud registration~\cite{colored-pcr}, eliminating the camera-wearer's movement. 
Fourth, a rotation sequence is obtained by computing the transformation between consecutive object point clouds using singular value decomposition. 
Combining these steps yields a sequence of 6DoF poses 
\(\boldsymbol{\tau} = \{\boldsymbol{\tau}_1, \boldsymbol{\tau}_2, \dots, \boldsymbol{\tau}_T\}\), 
where \(\boldsymbol{\tau}_t = (x, y, z, \mathrm{roll}, \mathrm{pitch}, \mathrm{yaw})\).
Here, $(x,y,z)$ captures the translational components of the object's centroid position, and $(\mathrm{roll}, \mathrm{pitch}, \mathrm{yaw})$ represents the rotational components of the object.
These trajectories represent the object’s pose over time, enabling us to approximate the end-effector states during manipulation without gripper states. 
As illustrated in Fig.~\ref{fig:egoscaler-example}, we successfully obtain action trajectories across diverse environments.

\myparagraph{Egocentric Video Resources.}
EgoScaler can be applied to various types of egocentric videos.
The original paper of EgoScaler targets only Exo-Ego4D~\cite{egoexo4d}, but this work expands it to four large egocentric video datasets with diverse activities and scenarios: Ego4D~\cite{ego4d}, Ego-Exo4D~\cite{egoexo4d}, HD-EPIC~\cite{hd-epic}, and Nymeria~\cite{nymeria}.
Ego4D and Ego-Exo4D have a broader range of scenarios, including cooking, experiments, crafting, and repair.
HD-EPIC primarily captures human activities in the cooking domain. 
Nymeria focuses on full-body motion understanding, resulting in fewer scenarios involving hand–object interactions.  
Unlike the other datasets, Ego4D does not provide camera intrinsics required by EgoScaler to reconstruct 3D trajectories.
We therefore estimate them in advance using COLMAP~\cite{sfm, mvs}.
COLMAP searches for an initial image pair that satisfies predefined feature-matching and geometric constraints, such as a minimum number of inliers and a maximum re-projection error.
If no valid pair is found, we exclude the corresponding instance from our dataset.  
By applying EgoScaler to these datasets, we initially obtained 124,559 episodes.

\myparagraph{Data Curation Methods.}
We found that EgoScaler sometimes produces inaccurate trajectories, mainly due to object detection and point cloud registration errors.
To remove them automatically, we apply two rule-based filters: a travel distance threshold for registration errors and a background track similarity threshold for detection errors.

For the travel distance threshold ($\delta_{DT}$), we define the travel distance $D$ of a trajectory as 
the cumulative displacement of its positional component, 
$D = \sum_{t=1}^{T-1} \lVert \mathbf{p}_{t+1} - \mathbf{p}_t \rVert_2$, 
where $\mathbf{p}_t=(x_t,y_t,z_t)$ denotes the translational element of the object trajectory $\boldsymbol{\tau}_t$. 
Trajectories corrupted by registration errors often contain abrupt mismatches across consecutive frames, leading to abnormally large $D$. 
We therefore discard trajectories with $D > \delta_{\text{TD}}$.

\begin{figure}
    \centering
    \includegraphics[width=0.95\linewidth]{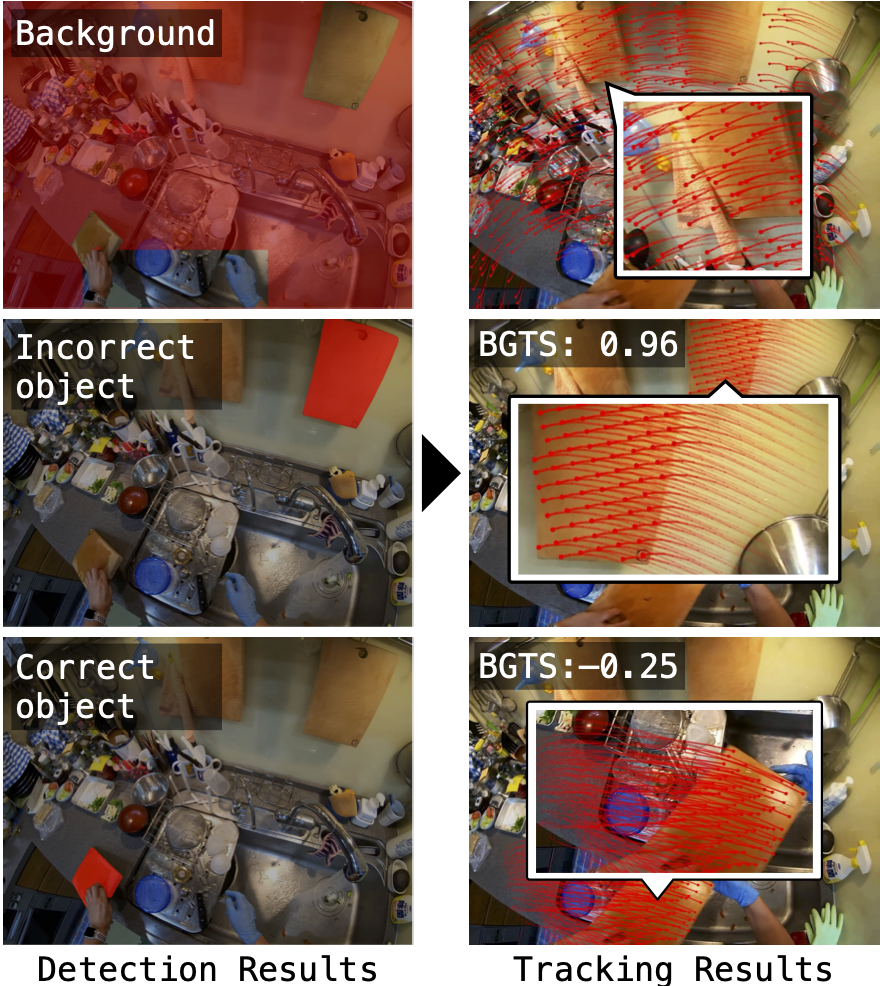}
    \caption{\textbf{Samples of computing background track similarity (BGTS).} Tracked sequences are depicted in red.
    Low BGTS indicates the object moves due to hand interaction.
    }
    \vspace*{-3pt}
    \label{fig:filter-example}
\end{figure}

For the background track similarity threshold (\(\delta_{\text{BGTS}}\)), 
let the object track be $\{\mathbf{o}_t\}_{t=1}^{T}$ and the background track be $\{\mathbf{q}_t\}_{t=1}^{T}$, 
where $\mathbf{o}_t,\mathbf{q}_t \in \mathbb{R}^2$ are image-plane positions obtained by point tracker~\cite{spatracker} within EgoScaler framework.
We observed that tracks from detection errors often resemble those of the background, as shown in Fig.~\ref{fig:filter-example}.
This is because detection errors typically occur on non-interacted, static objects. 
To detect such cases, we compute the background track similarity (BGTS) as the average cosine similarity between the object and background displacements:
\begin{equation}
    \text{BGTS} = \frac{1}{T-1} \sum_{t=1}^{T-1} 
    \frac{\mathbf{u}_t \cdot \mathbf{b}_t}{\|\mathbf{u}_t\| \, \|\mathbf{b}_t\|},
\end{equation}
where $\mathbf{u}_t = \mathbf{o}_{t+1}-\mathbf{o}_t$ and $\mathbf{b}_t = \mathbf{q}_{t+1}-\mathbf{q}_t$ are velocity vectors from the object and background tracks. 
We discard episodes with $\text{BGTS} > \delta_{\text{BGTS}}$, where we empirically set $\delta_{\text{BGTS}}=0.7$ based on simulator experiments.

Moreover, due to depth inconsistencies between consecutive frames, the translational components of the extracted trajectories often contain jitter noise.  
To suppress this, we apply a smoothing filter by averaging each translation vector over a five-frame window centered at the current frame.  
At sequence boundaries (i.e., \(t = 1, 2, T{-}1, T\)), the window size is reduced accordingly, increasing the influence of the central frame.  
Applying these curation methods, we finally obtain 45,157 episodes.  
The statistics of our dataset, along with teleoperation-based robot datasets, are shown in Table~\ref{tab:dataset-stats}.

\begin{figure}
    \centering
    \includegraphics[width=0.94\linewidth]{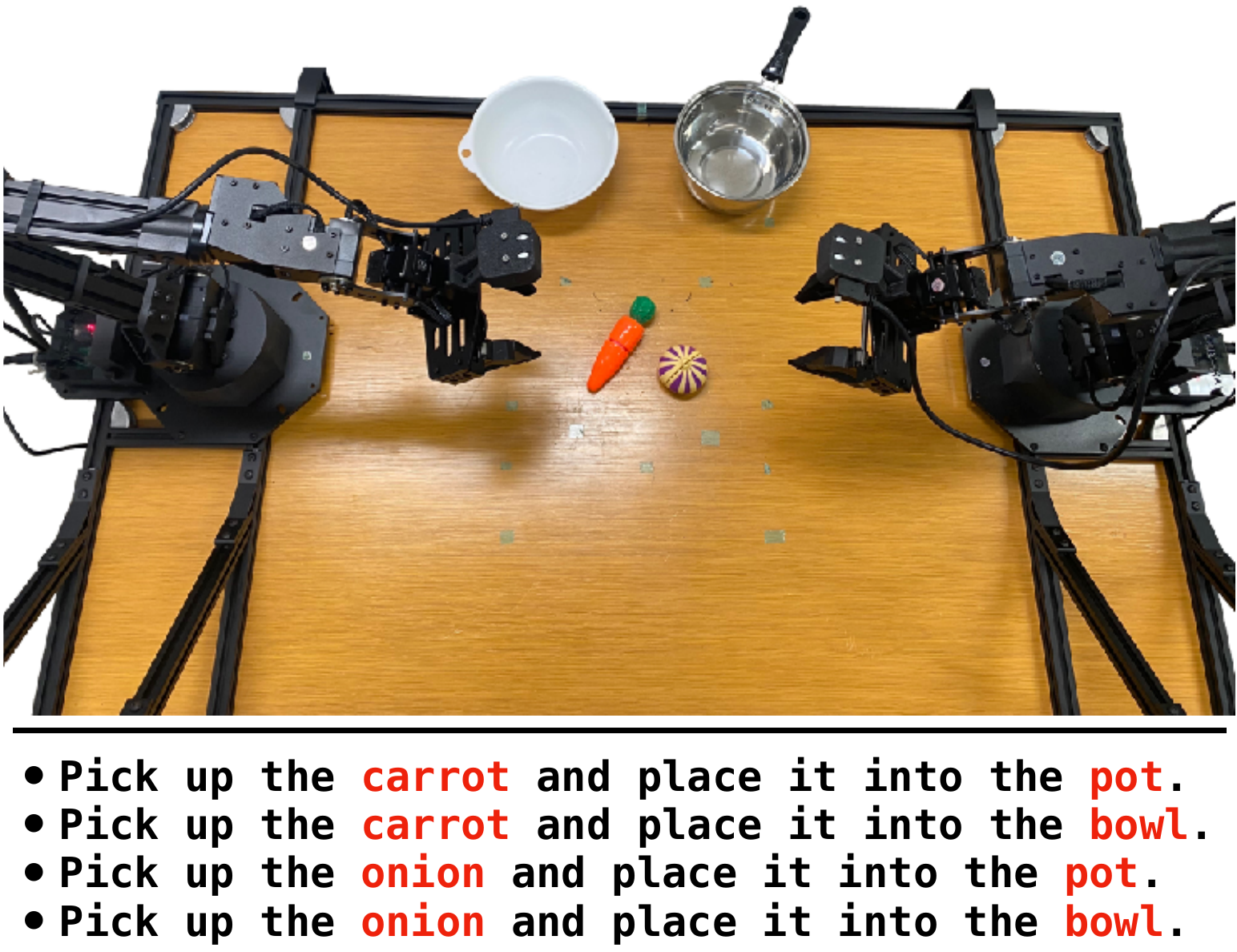}
    \vspace*{-3pt}
    \caption{\textbf{Overview of real-robot evaluation setting.}
    }
    \label{fig:exp-overview}
\end{figure}
\begin{table}[t]
    \centering
    \footnotesize
    \caption{\textbf{Statistics of previous robot datasets and ours.}}
    \vspace*{-3pt}
    \begin{tabular}{lrrrr}
    \toprule
        Dataset & \#Episodes & \#Verbs & \#Objects \\
    \midrule
        RoboTurk~\cite{roboturk} & 1,796 & 2 & 2 \\
        BC-Z~\cite{bc-z} & 39,350 & 9 & 17 \\
        BridgeData V2~\cite{bridge} & 53,192 & 270 & 749 \\
        Fractal~\cite{rt-1} & 87,212 & 6 & 13 \\
        DROID~\cite{droid} & 92,233 & 194 & 907 \\
    \midrule
        Ours & 45,157 & 313 & 1,217 \\
    \bottomrule
    \end{tabular}
    \label{tab:dataset-stats}
\end{table}

\subsection{Policy Training}
In this work, we use a state-of-the-art VLA $\pi_0$~\cite{pi0}.
Built on the pre-trained VLM PaliGemma~\cite{paligemma}, $\pi_0$ employs a flow-matching–based~\cite{flow-matching-gen-model, rectified-flow} action head that incorporates robot state and generates continuous, high-frequency actions.

\begin{figure*}[t]
    \centering
    \includegraphics[width=0.96\linewidth]{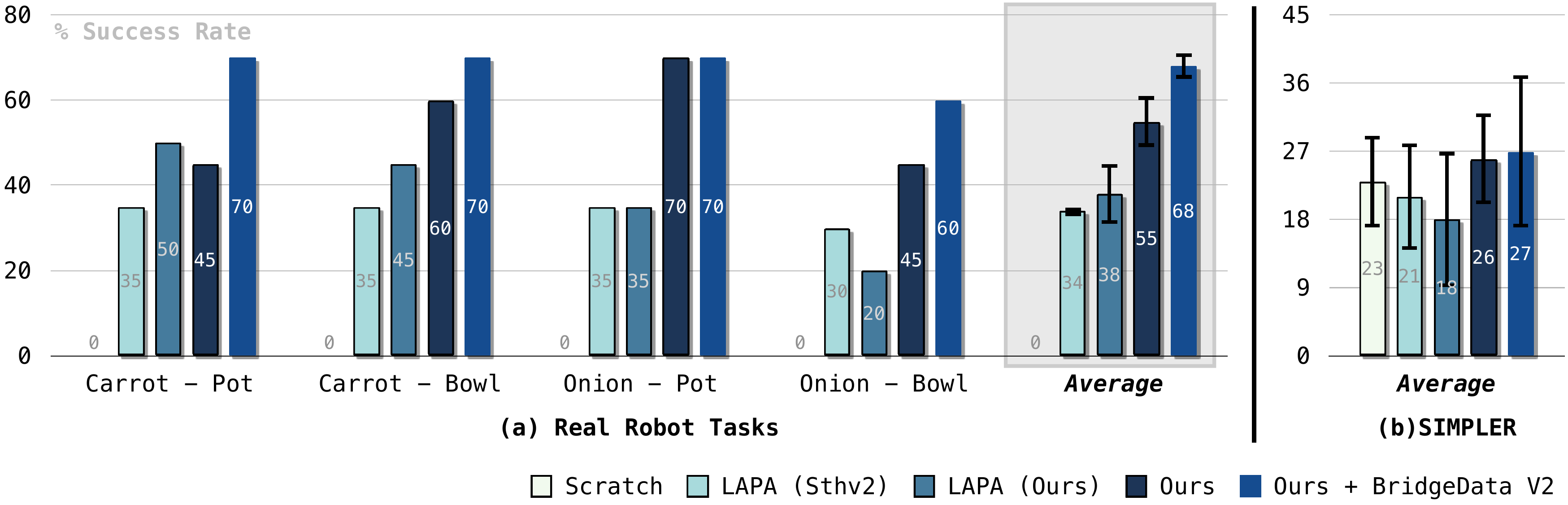}
    \vspace*{-3pt}
    \caption{\textbf{Performance of baselines and our dataset across manipulation tasks.}
    ``\texttt{ObjectA} – \texttt{ObjectB}'' denotes the task ``\texttt{pick ObjectA and place it into ObjectB}.''
    Parentheses indicate the used dataset for latent action pre-training.
    Average success rates (\%) with standard error are reported.
    }
    \label{fig:main-results}
\end{figure*}

\myparagraph{Action Representation.}
During pre-training on our dataset, an action is represented as the displacement of the 6DoF object pose trajectory~$\boldsymbol{\tau}$:
\begin{equation}
\mathbf{a}_t = \bigl[\Delta x_t,\;\Delta y_t,\;\Delta z_t,\;\Delta\mathrm{rot6D}_t\bigr].
\end{equation}
Here, $\{x_t, y_t, z_t\}$ denotes the positional coordinates, and $\text{rot6D}_t \in \mathbb{R}^6$ is a flattened vector of the first two columns of the rotation matrix $R_t$.
Because gripper states cannot be obtained from Section \ref{subsec:dataset-construction}, each action is represented by a 9-dimensional vector. For proprioceptive states, we use the original trajectories $\boldsymbol{\tau}$ for each timestep, converting rotational elements into $\mathrm{rot6D}$ representation.
To mitigate distribution differences between human and robot data, all actions and proprioceptive states are normalized during training~\cite{egomimic}.

\myparagraph{Dataset Merging for Pre-training.}
Pre-training VLAs often involves combining multiple robot-embodiment datasets to enable large-scale and diverse training~\cite{oxe, pi0}. 
Following these, when merging our dataset with existing robot datasets, we pad and normalize action and proprioceptive
vectors as needed to match dimensionalities across datasets, and then perform joint training on the concatenated data.

\myparagraph{Training Objective.}
The model is trained to predict a sequence of future actions conditioned on language, vision, and proprioception inputs.  
We minimize the mean squared error between the predicted action \(\hat{\mathbf{a}}_t\) and the ground truth action \(\mathbf{a}_t\) across a chunk of \(H\) future steps:
\begin{equation}
\mathcal{L}_{\text{action}} = \frac{1}{H} \sum_{t=1}^{H} \left\| \mathbf{\hat{a}}_t - \mathbf{a}_t \right\|_2^2.
\end{equation}
This loss is used for both pre-training and post-training.

\section{Experiments}

We evaluate the effectiveness of our dataset as a pre-training source in simulated and real-robot environments.
For all experiments, we employ the $\pi_0$~\cite{pi0} architecture.
Unlike the conventional settings of fine-tuning a publicly available $\pi_0$ checkpoint, we \textbf{pre-train} a $\pi_0$ model within our dataset.

\subsection{Experimental Setup}
\myparagraph{Manipulation Task Details.}
For simulated environments, we use SIMPLER~\cite{simplerenv} BridgeData V2 environment, which contains four pick-and-place tasks.
Inspired by~\cite{lapa}, we decided to collect a small amount of successful rollouts for the post-training purpose.
For this purpose, we applied a pre-trained $\pi_0$~\cite{pi0} on $\pi$~\cite{pi0} dataset and post-trained it on BridgeData V2~\cite{bridge} dataset.
Using the $\pi_0$, 25 successful rollouts were collected for each task. 
For evaluation, performance is measured with 200 rollouts for each task.

For real-robot environments, as shown in Fig.~\ref{fig:exp-overview}, we use ALOHA~\cite{aloha} and design four language-aware pick-and-place tasks with four objects. 
The task instruction follows the template: ``\texttt{Pick up the [carrot/onion] and place it into the [pot/bowl]}.''
In each rollout, all four objects are present simultaneously, requiring the model to interpret the given instruction correctly.
We manually collect 200 episodes per task, totaling 800 episodes.  
For evaluation, performance is measured with 10 rollouts for each task.
The success rate is calculated using a two-stage scoring scheme: 0.5 points for grasping the correct object, and an additional 0.5 points for placing it in the correct location.

\myparagraph{Implementation Details.}
We trained \(\pi_0\) on 8×H200 GPUs using AdamW~\cite{adamw} optimizer with bfloat16 precision under a constant learning rate of \(5\times10^{-5}\).
We freeze the pre-trained VLM parameters during both pre-training and post-training. 
This design, inspired by SmolVLA~\cite{smolvla}, makes training GPU-friendly and time-efficient while preserving competitive performance.
Pre-training was conducted for 20,000 steps with a batch size of 1,024. 
In evaluation, we use 40,000 steps with a batch size of 128 for the real-robot setting. 
For the simulator setting, we select the best checkpoint on a validation set, evaluated every 10,000 steps between 10,000 and 50,000 steps.

\subsection{Results}\label{subsec:results}

\myparagraph{Comparison with Scratch and Our Approach.}
The scratch baseline is a model trained only on the post-training dataset.
As shown in Fig.~\ref{fig:main-results}, our method outperforms the scratch baseline in real-robot tasks, while achieving smaller but consistent gains in simulation. 
Under an identical architecture and post-training setting, the scratch model attains 0\% success in real-robot, indicating a failure to ground instructions and generate meaningful trajectories. 
The smaller performance gap in simulation is likely due to a visual domain gap, which we discuss in Section~\ref{sec:ablation}.
Moreover, merging our dataset with BridgeData V2~\cite{bridge} yields additional improvements compared to our dataset alone.

\myparagraph{Comparison of LAPA and Our Approach.}  
LAPA~\cite{lapa} is an implicit pre-training approach that does not rely on auxiliary labels, enabling the use of egocentric videos as VLA pre-training data.  
In latent action pre-training, a discrete set of ``action'' tokens is first learned 
via a VQ-VAE~\cite{vqvae} quantizer applied to temporally separated image frame pairs.  
This approach then pre-trains a VLA to predict these tokens from image–text pairs.
We apply the pretraining methodology of LAPA to the same $\pi_0$ model instead of the original VLA architecture proposed in the paper for the sake of fair comparison.
For the latent action pre-training setting, we conduct experiments on both our dataset without action labels and Something-Something V2 (SthV2)~\cite{sthv2} dataset used in the paper.  
SthV2 consists of short first- and third-person perspective videos, covering human actions with a variety of objects. 

Fig.~\ref{fig:main-results} presents the performance comparison between LAPA and ours. 
Our approach consistently outperforms LAPA in both real-robot and simulated environments. 
Although LAPA outperforms the scratch baseline in real-robot tasks, its performance degrades in simulation, likely due to the simplified visual systems in simulated environments. 
This suggests that leveraging rich, diverse egocentric video without action labels can harm performance in simulation.
Similarly, the greater visual diversity in our dataset is more effective than SthV2 in the real-robot setting, but it leads to performance drops in simulation.
These results indicate that explicit action trajectories provide more robust and informative supervision across environments.

\begin{table}[t]
    \centering
    \footnotesize
    \caption{\textbf{Comparison with robot datasets.} 
    Successes out of 10 rollouts are reported for each task, with the final column showing the total.
    }
    \vspace*{-3pt}
    \resizebox{\linewidth}{!}{
    \begin{tabular}{lccccc}
    \toprule
    Dataset & \makecell{\texttt{Carrot}\\\texttt{-Pot}} 
            & \makecell{\texttt{Carrot}\\\texttt{-Bowl}} 
            & \makecell{\texttt{Onion}\\\texttt{-Pot}} 
            & \makecell{\texttt{Onion}\\\texttt{-Bowl}} 
            & \textit{Total} \\
    \midrule
    BridgeData V2~\cite{bridge} & 4/10 & 3/10 & 6/10 & 4/10 & 17/40 \\
    BC-Z~\cite{bc-z}            & 5/10 & 5/10 & 4/10 & 5/10 & 19/40 \\
    Fractal~\cite{rt-1}            & 7/10 & 4/10 & 7/10 & 4/10 & 22/40 \\
    \midrule
    Ours                        & 4/10 & 6/10 & 7/10 & 4/10 & 21/40 \\
    Ours + \cite{bridge}        & 7/10 & 7/10 & 7/10 & 6/10 & \textbf{27/40} \\
    \bottomrule
    \end{tabular}
    }
    \label{tab:robodata-comparison}
\end{table}

\myparagraph{Comparison of Robot Datasets and Ours.}\label{sec:robotdata-comparison}
Table~\ref{tab:robodata-comparison} summarizes the performance of $\pi_0$ pre-trained separately on three robot datasets (Fractal~\cite{rt-1}, BridgeData V2~\cite{bridge}, and BC-Z~\cite{bc-z}) and on our dataset in real-robot tasks.
This comparison reveals two key findings.
First, merging ours with BridgeData V2 outperforms using any robot dataset alone. 
This result suggests that our dataset can effectively complement robot datasets and serve as a useful component within larger multi-embodiment collections such as OXE~\cite{oxe}.
Second, pre-training on our dataset alone yields higher performance than BC-Z and BridgeData V2, but lower than Fractal, owing to its much larger scale.
Together, these results highlight the importance of large-scale pre-training for VLAs and demonstrate that our dataset is effective both on its own and when combined with existing robot datasets.

\subsection{Ablation Study}\label{sec:ablation}

\begin{figure}[t]
    \centering
    \includegraphics[width=0.94\linewidth]{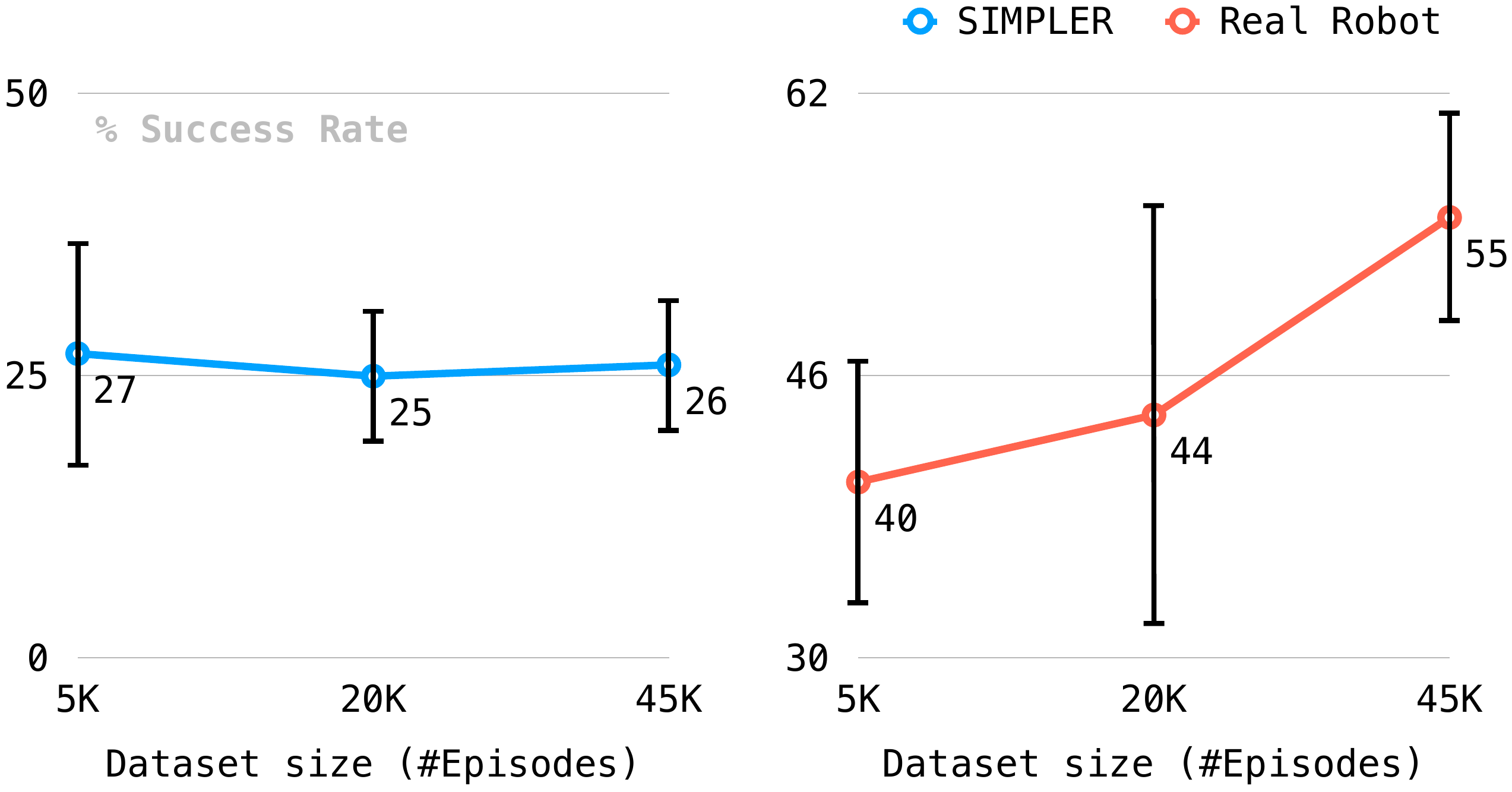}
    \vspace*{-2pt}
    \caption{\textbf{Dataset size and performance.}
    Average success rates (\%) with standard error reported.}
    \label{fig:scale-result}
\end{figure}

\myparagraph{Dataset Scalability.}
As our framework automatically extracts object manipulation trajectories, we examine how dataset scale affects performance.  
We utilize data at ratios of 1.0, 0.5, and 0.1 of the full dataset (corresponding to 45K, 20K, and 5K episodes, respectively).
Fig.~\ref{fig:scale-result} illustrates the results across different dataset sizes. In the real-robot setting, scaling our dataset significantly improves task performance.  
In contrast, in simulation, the full dataset slightly underperforms the 0.1-ratio subset by 1\% on average. 
The limited improvement from scaling in simulation likely stems from a visual domain gap. 
While the simulator provides reduced noise and controlled variability, the rich and diverse cues in egocentric videos may hinder performance.

\myparagraph{Hyperparameter of Background Trajectory Similarity.}
Setting an appropriate curation threshold is crucial to balancing the scale and quality of our dataset.  
We conduct an ablation study, varying the background trajectory similarity threshold $\delta_{\text{BGTS}} = \{0.5, 0.7, 1.0\}$.  
As shown in Table~\ref{tab:ablate-thresh}, a lower threshold ($\delta_{\text{BGTS}} = 0.5$) removes noisy trajectories but significantly reduces the dataset scale.  
In contrast, a higher threshold ($\delta_{\text{BGTS}} = 1.0$) retains more episodes but leads to degraded performance due to increased noise.  
We find that $\delta_{\text{BGTS}} = 0.7$ achieves the best trade-off between dataset size and task performance in simulation experiments. 
We therefore adopt this value for constructing our dataset.

\begin{table}[t]
    \centering
    \small
    \vspace*{-3pt}
    \caption{\textbf{Performance comparison across different background track similarity thresholds (BGTS).}
    }
    \vspace*{-3pt}
    \begin{tabular}{lrrr}
    \toprule
        $\delta_{\text{BGTS}}$ & \#Episodes & SIMPLER~\cite{simplerenv} & Real Robot \\
    \midrule
        0.5 & 28,719 & 25.8$_{\pm 5.4}$ & 53.8$_{\pm 5.2}$ \\
        0.7 & 45,157 & \textbf{26.0$_{\pm 5.9} $} & \textbf{55.0$_{\pm 6.1}$} \\
        1.0 & 86,427 & 22.5$_{\pm 4.8}$ & 38.8$_{\pm 4.3}$ \\ 
    \bottomrule    
    \end{tabular}
    \label{tab:ablate-thresh}
\end{table}
\section{Conclusion}
In this paper, we demonstrated that egocentric videos without auxiliary labels can serve as an effective resource for VLA pre-training. 
Using EgoScaler, we constructed a large-scale dataset by extracting explicit object manipulation trajectories from egocentric videos. 
Pre-training on this dataset yields performance competitive with real-robot datasets and significantly outperforms training from scratch and LAPA.
Moreover, merging our dataset with a real-robot dataset further boosts performance.
These findings highlight that utilizing egocentric videos is a promising step toward addressing the data scarcity problem in robot learning. 



\section*{ACKNOWLEDGMENT}
This work was supported by JSPS KAKENHI Grant Number JP22K17983, JP22KK0184 and JST CRONOS JPMJCS24K6.



\bibliographystyle{IEEEtran}
\bibliography{reference}

\end{document}